\documentclass[10pt,twocolumn,letterpaper]{article}

\usepackage{iccv}
\usepackage{times}
\usepackage{epsfig}
\usepackage{graphicx}
\usepackage{subfig}
\usepackage{amsmath}
\usepackage{amssymb}
\usepackage{authblk}
\usepackage[titletoc,toc,title]{appendix}

\usepackage[breaklinks=true,bookmarks=false]{hyperref}

\iccvfinalcopy 

\def\iccvPaperID{1275} 

\ificcvfinal\fi
\begin{document}

\title{DeepVisage: Making face recognition simple yet with powerful generalization skills}


\author[1]{Abul Hasnat}
\author[2]{Julien Bohn\'{e}}
\author[2]{Jonathan Milgram}
\author[2]{St\'{e}phane Gentric}
\author[1]{Liming Chen}
\affil[1]{Laboratoire LIRIS, \'{E}cole centrale de Lyon, 69134 Ecully, France.}
\affil[2]{Safran Identity \& Security, 92130 Issy-les-Moulineaux, France.}
\affil[ ]{{md-abul.hasnat@ec-lyon.fr, \{julien.bohne, stephane.gentric, jonathan.milgram\}@safrangroup.com, liming.chen@ec-lyon.fr}}


\maketitle

\begin{abstract}
   Face recognition (FR) methods report significant performance by adopting the convolutional neural network (CNN) based learning methods. Although CNNs are mostly trained by optimizing the softmax loss, the recent trend shows an improvement of accuracy with different strategies, such as task-specific CNN learning with different loss functions, fine-tuning on target dataset, metric learning and concatenating features from multiple CNNs. Incorporating these tasks obviously requires additional efforts. Moreover, it demotivates the discovery of efficient CNN models for FR which are trained only with identity labels. We focus on this fact and propose an easily trainable and single CNN based FR method. Our CNN model exploits the residual learning framework. 
Additionally, it uses normalized features to compute the loss.
Our extensive experiments show excellent generalization on different datasets. We obtain very competitive and state-of-the-art results on the LFW, IJB-A, YouTube faces and CACD datasets.
\end{abstract}

\section{Introduction}
\label{sec:intro}
Face recognition (FR) is one of the most demanding computer vision tasks, due to its practical use in numerous applications, such as biometric, surveillance and human-machine interaction.
The state-of-the-art FR methods \cite{taigman2014deepface, schroff2015facenet, sun2015deepid3, parkhi2015deep, baiduliu2015} surpassed human performance (97.53\%) and achieved significant accuracy on the standard labeled faces in the wild (LFW) \cite{lfw_huang2014} benchmark.
These remarkable results are achieved by training the deep convolutional neural network (CNN) \cite{recentcnngu2015} with large databases \cite{mscelebguo16, parkhi2015deep, yi2014learning, umdfacesbansal2016}.

The facial image databases mostly provide the identity labels. These labels allow the CNN models to be easily trained with the softmax loss. 
FR methods generally use the trained CNN model to extract facial features and then perform verification by computing distance or recognition with a classifier. 
However, from our extensive study (see Sect. \ref{sec:rel_work}), we observe that recent methods include different additional strategies to obtain better performance, such as: 
\begin{enumerate}
\item \textit{train CNN with different loss functions \cite{schroff2015facenet, sun2015deepid3}:} requires carefully preparing the image pairs/triplets by maintaining certain constrains \cite{schroff2015facenet}, because arbitrary pairs/triplets do not contribute to the training. Online triplet generation requires a larger batch size (e.g., \cite{schroff2015facenet} used 1.8K images in a mini-batch with 40 images/identity), which is excessive for a limited resource machine. On the other hand, using offline triplets can be critical as many of them will be useless while training progresses. The joint optimization \cite{sun2015deepid3} with Softmax and Contrastive losses not only requires specific training data (with identity and pair labels) but also complicates the training procedure.
\item \textit{fine-tune CNN:} requires training on each target dataset, which restricts the ability to generalize.
\item \textit{metric learning \cite{Sankaranarayanan2016a, ding2015robust}:} requires particular form of training data (e.g., triplets). Moreover, it does not always guarantee to enhance performance \cite{wang2016face}.
\item \textit{concatenating features from multiple CNNs \cite{sun2015deepid3, baiduliu2015}:} requires additional training data of different forms and train CNNs for each form. Besides, it is necessary to explore the particular modalities that can contribute to enhance performance.
\end{enumerate}
%

The use of the above strategies requires significant efforts in terms of data preparation or selection and computing resources. 
On the other hand, recent results on the ImageNet challenge \cite{imagenet_ijcv} indicate that deeper CNNs enhance performance of different computer vision tasks.
These observations raise the following question - \textit{can we achieve state-of-the-art results with a single CNN model which is trained only once with the identity labels?} Our research is motivated by this question and we aim to address it by developing a simple yet robust single-CNN based FR method. Moreover, we want that our once-trained single CNN based FR method generalizes well across different datasets.
%

In this research, our primary objective is to discover an efficient CNN architecture. 
We follow the recent findings, which suggest that deeper CNNs perform better on a number of computer vision tasks \cite{resnethe2015deep, recentcnngu2015}.
We construct a deep CNN model with 27 convolutional and 1 fully connected (FC) layers, which incorporates the residual learning framework \cite{resnethe2015deep}. 
Moreover, we aim to find an efficient way to train our CNN only with the identity labels. 
%
Recently, \cite{centerlosswen2016} achieves high FR performance with a CNN trained from the identity labels. However, they perform joint optimization using the softmax and center loss \cite{centerlosswen2016} (CL). CL improves the features discrimination among different classes. 
It follows the principle that, features learned from a deep CNN should minimize the intra-class distances. 
Interestingly, we observe (see Fig \ref{fig:comp_bn_cl}) that an equivalent representation can be achieved by normalizing the CNN features before computing the loss. Therefore, we train our CNN using the softmax loss with the normalized features. 

With our single CNN model, first we evaluate on the LFW \cite{lfw_huang2014} benchmark and observe that it obtains  99.62\% accuracy. In order to demonstrate its effectiveness, we evaluated it on different challenging face verification tasks, such as face templates matching on the IJB-A \cite{ijbaKlare2015} dataset, video faces matching on the YouTube Faces \cite{ytfwolf2011} (YTF) dataset and cross age face matching on the CACD \cite{chen2015facecacd} dataset.
Our method achieves 82.4\%  TAR@FAR=0.001 on IJB-A \cite{ijbaKlare2015}, 96.24\% accuracy on YTF \cite{ytfwolf2011} and 99.13\% accuracy on CACD \cite{chen2015facecacd}. These results indicate that our method achieves very competitive and state-of-the-art results. Moreover, it generalizes very well across different datasets.

We summarize our contributions as follows: (a) conduct extensive study and provide (Sec \ref{sec:rel_work}) a review and methodological comparison of the state-of-the-art methods; (b) propose (Sect. \ref{sec:deepvisage}) an efficient single CNN based FR method; (c) conduct (Sect. \ref{sec:res_exp}) extensive experiments on different datasets, which demonstrate that our method has excellent generalization ability; and (d) perform (Sect. \ref{ssec:disscussion}) an in-depth analysis to identify the influences of different aspects.
%

In the remaining part of this paper, first we study and analyze the state-of-the-art FR methods in Section \ref{sec:rel_work}, describe our proposed method in Section \ref{sec:deepvisage}, present experimental results, perform analysis of our method and discuss them in Section \ref{sec:res_exp} and finally draw conclusions in Section \ref{sec:conclusion}.
\section{Related work, state-of-the-art FR methods}
\label{sec:rel_work}
Face recognition (FR) in unconstrained environment attracts significant interest from the community. Recent methods exploited deep CNN models and achieved remarkable results on the LFW \cite{lfw_huang2014} benchmark.
Besides, numerous methods have been evaluated on the IJB-A \cite{ijbaKlare2015} dataset. We study\footnote{We consider only the CNN based methods. For the others, we refer readers to the recently published survey \cite{learned2015labeled} for LFW and \cite{ijbaKlare2015} for IJB-A.} and analyze these methods based on several key aspects: (a) details of the CNN model; (b) loss functions used; (c) incorporation of additional learning strategy; (d) number of CNNs and (e) the training database used.

Recent methods tend to learn CNN based features using a \textit{deep architecture} (e.g., 10 or more layers). This is inspired from the extraordinary success on the ImageNet \cite{imagenet_ijcv} challenge by famous CNN architectures \cite{recentcnngu2015}, such as AlexNet, VGGNet, GoogleNet, ResNet, etc. The FR methods commonly use these architectures as their baseline model (directly or slightly modified). For example, AlexNet is used by \cite{Sankaranarayanan2016, Sankaranarayanan2016a, viplfacenetliu2016, allinoneranjan2016, abdalmageed2016face, masi2016pose, schroff2015facenet}, VGGNet is used by \cite{parkhi2015deep, Crosswhite2016, masi16dowe, abdalmageed2016face, masi2016pose, ding2015robust, sparsifyingsun2015} and GoogleNet is used by \cite{nanyang2016, schroff2015facenet}. 
CASIA-Webface \cite{yi2014learning} proposed a simpler CNN model, which is used by \cite{wang2016face, chen2016unconstrained, ding2015robust}. Several methods, such as \cite{deepid2psun2015, taigman2014deepface, webscaletaigman2015, centerlosswen2016, sun2015deepid3} use a model with lower depth but increase its complexity with locally connected convolutional layers. Besides, \cite{zhou2015naive} use 4 parallel 10 layers CNNs to learn features from different facial regions. \textit{We follow the ResNet \cite{resnethe2015deep} based deep CNN model}.

FR methods often train multiple CNNs and accumulate features from all of them to construct the final facial descriptors. It provides an additional boost to the performance. Different types of inputs are used to train these multiple CNNs: (a) \cite{deepid2psun2015, sun2015deepid3, sparsifyingsun2015, wang2016face, ding2015robust, baiduliu2015} used image-crops focused on certain facial regions (eyes, nose, lips, etc.); (b) \cite{ding2015robust, abdalmageed2016face, masi2016pose, taigman2014deepface} used different modality of input images, such as 2D, 3D, frontalized and synthesized faces at different poses and (c) \cite{webscaletaigman2015, baiduliu2015} used different training databases with varying number of images. \textit{We do not follow this approach and train only one CNN}.

The CNN model parameters are learned by optimizing loss functions, which are defined based on the given task (e.g., classification, regression) and the available information (e.g., class labels, price). The softmax loss \cite{recentcnngu2015} is a common choice for classification tasks. It is often used by the FR methods to create good face representation by training the CNN as an identity classifier. It requires only the identity labels. The contrastive loss \cite{recentcnngu2015, contrastive_loss} is used by \cite{deepid2psun2015, taigman2014deepface, sparsifyingsun2015, sun2015deepid3, nanyang2016} for face verification and requires face image pairs and similarity labels. The triplet loss \cite{schroff2015facenet} is used by \cite{schroff2015facenet, parkhi2015deep, Crosswhite2016, baiduliu2015} for face verification and requires the face triplets. Recently the center loss \cite{centerlosswen2016} is proposed to enhance feature discrimination, which uses the identity labels. \textit{We use the softmax loss}. 

Several methods use multiple loss functions and train CNN using joint optimization \cite{deepid2psun2015, sparsifyingsun2015, sun2015deepid3, centerlosswen2016, allinoneranjan2016}. The other way is to use them sequentially \cite{taigman2014deepface, parkhi2015deep, Crosswhite2016, baiduliu2015, nanyang2016}, i.e., first train with the softmax and then train with the other loss. We observe that using multiple loss functions complicates the training data preparation task and the CNN training procedure. \textit{Therefore, we avoid this type of strategies}.

Fine-tuning the CNN parameters is a particular form of transfer learning. It is commonly employed by several methods \cite{wang2016face, chen2016unconstrained, Sankaranarayanan2016} on the IJB-A \cite{ijbaKlare2015} dataset. 
It refines the CNN parameters from a previously learned model using a target specific training dataset. 
Several methods do not directly use the raw CNN features but employ an additional learning strategy. 
The \textit{metric/distance learning} strategy based on the Joint Bayesian method \cite{chen2012bayesian} is a popular one and used by \cite{deepid2psun2015, yi2014learning, wang2016face, chen2016unconstrained, sparsifyingsun2015, sun2015deepid3, ding2015robust}. Recently, two different strategies \cite{Sankaranarayanan2016a, Sankaranarayanan2016a} have been proposed to learn feature embedding using face triplets. Another strategy, called template adaptation \cite{Crosswhite2016}, exploits an additional SVM classifier. Apart from these, principal component analysis (PCA) is used by several methods \cite{masi16dowe, abdalmageed2016face, masi2016pose} to learn a dataset specific projection matrix.
\cite{nanyang2016} learns an aggregation module to compute scores among two videos. 
The above methods often need training data from the target datasets. Moreover, they \cite{Sankaranarayanan2016, Sankaranarayanan2016a} may need to carefully prepare the training data, e.g., triplets. \textit{We do not need any such learning strategies}.
The use of a large facial training dataset is important to achieve high FR accuracy \cite{schroff2015facenet, zhou2015naive}. \cite{zhou2015naive} provided an in-depth analysis and demonstrated the effect of the dataset size and the number of identities for FR.
Following the high demand of a large FR dataset, several publicly available datasets have been released recently. Among them, CASIA-WebFace \cite{yi2014learning} is used by numerous methods \cite{centerlosswen2016, Sankaranarayanan2016, Sankaranarayanan2016a, viplfacenetliu2016, allinoneranjan2016, yi2014learning, wang2016face, chen2016unconstrained, ding2015robust, wu2015lightened, masi16dowe, abdalmageed2016face, masi2016pose}. Several researches \cite{masi16dowe, abdalmageed2016face, masi2016pose} enlarge it by synthesizing facial images with different shapes and poses based on the 3D face models.
Recently, the MSCeleb \cite{mscelebguo16} dataset has been publicly released. It contains the largest collection of facial images and identities. \textit{We exploit it to develop our FR method}.
%
%
\section{Proposed Method}
\label{sec:deepvisage}
Our FR method, called \textit{DeepVisage}, consists in pre-processing face image, learning CNN based facial features and computing similarity.
Following the recent trend \cite{taigman2014deepface, schroff2015facenet, sun2015deepid3, parkhi2015deep, yi2014learning}, we exploit the CNN as the core component.
Our deep CNN model follows the residual learning framework \cite{resnethe2015deep}. Moreover, it intelligently exploits feature normalization, which is a crucial step, see Sect. \ref{ssec:disscussion}. 
Our pre-processing stage consists in the detection of the face and facial landmarks, which are used to create a normalized face image.
We compute the cosine similarity among the features of a pair of faces as the verification score. Below, we describe these elements.

\subsection{Building blocks and deep CNN architecture}
\label{ssec:deep_cnn}
\paragraph{Convolutional networks:}
\label{cnn_architecture}
We begin with the basic ideas of CNN \cite{lecun1998gradient}: (a) local receptive fields with identical weights via the convolution operation and (b) spatial sub-sampling via the pooling operation. At a particular layer  $l$, the convolution of the input $f_{x,y}^{Op, l-1}$ to obtain the $k^{th}$ output feature map $f_{x,y,k}^{C,l}$, can be expressed as:
\begin{equation}
\label{eq:conv_op}
f_{x,y,k}^{C, l} = {\mathbf{w}_k^l}^T f_{x,y}^{Op, l-1} + b_k^l
\end{equation}
where, $\mathbf{w}_k^l$ and $b_k^l$ are the shared weights and bias. $C$ denotes convolution and $Op$ (for $l>1$) denotes various tasks, such as convolution, sub-sampling or activation. For $l=1$, $Op$ represents the input image.
Sub-sampling or pooling performs a simple local operation, such as computing the average or maximum value in a local spatial neighborhood followed by reducing spatial resolution. We apply max pooling for our CNN, which has the following form:
\begin{equation}
\label{eq:max_pool}
f_{x,y,k}^{P, l} = \max_{(m,n) \in \mathcal{N}_{x,y} } f_{m,n,k}^{Op, l-1}
\end{equation}
where, $\mathcal{N}_{x,y}$ denotes the local spatial neighborhood of $(x,y)$ coordinate and $P$ denotes the pooling operation.

In order to ensure non-linearity of the network, the feature maps are passed through a non-linear activation function, e.g., the Rectified Linear Unit (ReLU) \cite{recentcnngu2015, prelu_he2015}: $f_{x,y,k}^l = max(f_{x,y,k}^{l-1}, 0)$. We apply the Parametric Rectified Linear Unit (PReLU) \cite{prelu_he2015} as the activation function, which has the following form:
\begin{equation}
\label{eq:prelu}
f_{x,y,k}^{A,l} = max(f_{x,y,k}^{Op, l-1}, 0) + \lambda_k min(f_{x,y,k}^{Op, l-1}, 0)
\end{equation}
where, $\lambda_k$ is a trainable parameter to control the slope of the linear function for the negative input values and $A$ denotes activation operation.

At the basic level, a CNN is constructed by stacking series of convolution, activation and pooling layers, see LeNet-5 \cite{lecun1998gradient} for an example. Often a layer with full connections is placed at the end of the stacked layers, called the fully connected (FC) layer. It takes all points (neurons) from the previous layer as input and connects it to all points (neurons) of the output layer.
%
%

\paragraph{Residual learning framework \cite{resnethe2015deep}:}
\label{residual_learning}
A recent trend \cite{recentcnngu2015} on the ImageNet \cite{imagenet_ijcv} challenge shows that deeper CNNs achieve better results. However, it increases the model complexity, which makes it harder to optimize the loss of the CNN model. Besides, they may generate higher training error than a shallower CNN \cite{resnethe2015deep}. 
The residual learning framework \cite{resnethe2015deep} provides a solution to these problems.

For a stack of a few layers, residual learning fits a mapping $\mathcal{F}(f) := \mathcal{H}(f) - f$ instead of fitting the underlying mapping $\mathcal{H}(f)$. Therefore, the original mapping is formulated as $\mathcal{F}(f) + f$, which means directly adding the input feature map $f$ with the output of the stacked layers $\mathcal{F}(f)$. This idea can be easily implemented with the notion of \textit{shortcut connection}. Formally, the output of a residual block $R$ can be expressed as:
\begin{equation}
\label{eq:res_block_op}
f_{x,y,k}^{R, l} = f_{x,y}^{Op, l-q} + \mathcal{F}(f_{x,y}^{Op, l-q}, \{W_k\})
\end{equation}
where, $f_{x,y}^{Op, l-q}$ represents the input feature map, $\mathcal{F}(.)$ is the residual mapping to be learned, $W_k$ is the parameters of the $k^{th}$ residual block and $q$ is the total number of stacked layers within the residual block. 
The flexible form of the residual function $\mathcal{F}(.)$  allows to stack multiple layers with different types of operations, such as convolution, pooling, activation etc. All of the residual blocks in our CNN consist of two convolution layers with different numbers of neurons. Each convolution is followed by a PReLU activation.
\paragraph{Loss function:}
\label{loss_function}
Deep CNNs are trained by optimizing loss function. We use the softmax loss, which is widely used for classification: 
\begin{equation}
\label{eq:softmax}
\mathcal{L}_{Softmax} = -\sum_{i=1}^{N}log \frac{e^{\mathbf{w}^T_{y_i}f_i + b_{y_i}}}{\sum_{j=1}^{K}e^{\mathbf{w}^T_jf_i + b_j}}
\end{equation}
where, $f_i$ and $y_i$ are the features and true class label of the $i^{th}$ image. $\mathbf{w}_j$ and $b_j$ denote the weights and bias of the $j^{th}$ class. $N$ and $K$ denote the number of training samples and the number of classes.

\paragraph{Feature normalization (FN):}
\label{feature_norm}
It is often used as a necessary step in many learning algorithms. It ensures that all of the features have equal contribution to the cost function \cite{PR_book}. With deep CNNs, we cannot guarantee this by only normalizing the input image pixels, because the scale of features (from the final FC layer) may change due to a series of operations at different layers. 
Therefore, to avoid the influence of un-normalized features during cost computation, we provide normalized features  $f^{Nr}_i$ to the softmax loss as:
$f^{Nr} = \frac{f^{Op} - \mu}{\sqrt{\sigma^2}}$, where $\mu$ and $\sigma^2$ are the mean and variance. 
%

\begin{figure}[t]
\centering
\hfil
\includegraphics[scale=0.16]{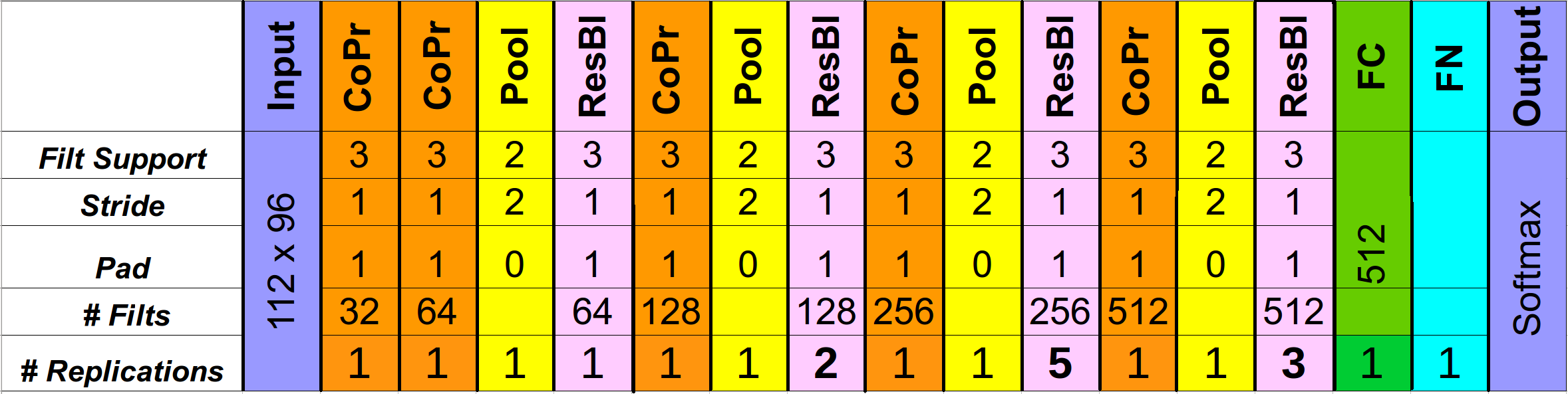}
\caption{\footnotesize{Illustration of the proposed CNN architecture. \textbf{\textit{CoPr}} indicates convolution followed by the PReLU activation function. \textbf{\textit{ResBl}} is a residual block which computes $output \; = \; input + CoPr(CoPr(input))$. \textit{\textbf{\# Replication}} indicates how many times the same block is sequentially replicated in the CNN model. \textit{\textbf{\# Filts}} denotes the number of feature maps. \textbf{\textit{FN}} denotes feature normalization.}}
\label{fig:res_block_cnn}
\end{figure}

During training, we apply normalization by computing $\mu$ and $\sigma$ from the samples of each mini-batch. Moreover, we maintain the moving average of $\mu$ and $\sigma$ and use them to normalize the test samples. Note that, this is a specific case of the popular batch normalization (BN) technique \cite{batch_normalization} with scale $\gamma=1$ and shift $\beta=0$.
\paragraph{Proposed CNN architecture:}
\label{proposed_cnn_model}
Our CNN model consists of 27 convolution (\textit{Conv}), 4 pooling (\textit{Pool}) and 1 fully connected (\textit{FC}) layers. Each convolution uses a $3\times3$ kernel 
and is followed by a PReLU activation function. The CNN progresses from the lower to higher depth by decreasing the spatial resolution using a $2\times2$ \textit{max Pool} layer while gradually increasing the number of feature maps from 32 to 512. We use a \textit{FC} layer of 512 neurons after the last \textit{Conv} layer. We normalize (see \textit{\textbf{FN}} above) the output of this \textit{FC} layer and consider it as the desired feature representation of the input image. Finally, we use the \textit{softmax} layer to compute the loss and optimize it during training. Our CNN model incorporates the residual learning framework \cite{resnethe2015deep}, see Fig. \ref{fig:res_block_cnn} for the details. Overall, it comprises 40.5M parameters.
\subsection{Image pre-processing and face verification}
\label{ssec:img_preprocess_face_verify}
\paragraph{\textit{Pre-processing:}}
\label{par:pre_processing}
We maintain the same form of the 2D face image during training and testing. Our pre-processing steps are: (a) detect\footnote{In case of multiple faces, we take the face closer to the image center.} face and landmarks using the MTCNN \cite{mtcnnZhang2016} detector; (b) 
normalize the face image by applying a 2D similarity transformation. The transformation parameters are computed from the location of the detected landmarks on the image and pre-set coordinates in a 112$\times$96 image frame; and (c) convert to grayscale.
\paragraph{\textit{Face verification:}}
\label{par:face_ver_test}
We verify a pair of face images \cite{lfw_huang2014}, templates \cite{ijbaKlare2015} (contains multiple images and video frames) and videos \cite{ytfwolf2011} (given as frames) using the following steps:
\begin{enumerate}
\item \textit{pre-process}: apply the pre-processing\footnote{If the landmarks detector fails we keep the face image by cropping it based on the given/detected bounding box.} stage described in the previous paragraph.
\item \textit{extract facial feature/representation}: we use the trained CNN model to extract the facial feature descriptor. For an image $i$, we obtain its descriptor $f_i$ by taking element-wise maximum of the features from its original $f_{i,o}$ and horizontally flipped version $f_{i,f}$. 
In order to perform verification based on template \cite{ijbaKlare2015} and video \cite{ytfwolf2011}, we obtain the descriptor for an identity by taking element-wise average of the features from all of the images/frames.
\item \textit{compute verification score}: for a given pair of facial features, we compute the cosine similarity as the verification score. 
We compare this score to a threshold to decide whether two images belong to the same person.
\end{enumerate}
\section{Experiments, Results and Discussion}
\label{sec:res_exp}
Our experiments consist of first training the CNN model and then use it to extract facial features and perform different types (single-image \cite{lfw_huang2014, chen2015facecacd}, multi-image or video \cite{ijbaKlare2015, ytfwolf2011}) of face verification. In order to verify the effectiveness, we experiment on several datasets, namely LFW \cite{lfw_huang2014}, IJB-A \cite{ijbaKlare2015}, YTF \cite{ytfwolf2011} and CACD \cite{chen2015facecacd}.

\subsection{CNN Training}
\label{ssec:cnn_train}
We collect the training images from the cleaned\footnote{We take the list of 5.05M faces provided by \cite{wu2015lightened} and keep non-overlapping (with test set) identities which has at least 30 images after successful landmarks detection.} version of the MS-Celeb-1M \cite{mscelebguo16} database, which consists of 4.47M images of 62.5K identities.
We train our CNN model using only the identity label of each image. We use 95\% images (4.2M images) for training and 5\% images (232K images) for monitoring and evaluating the loss and accuracy. We train our CNN using the \textit{stochastic gradient descent} method and \textit{momentum} set to 0.9. Moreover, we apply \textit{L2} regularization with the \textit{weight decay} set to $5e^{-4}$. We begin the CNN training with a learning rate 0.1 for 2 epochs. 
Then we decrease it after each epoch by a factor 10. We stop the training after 5 epochs.
We use 120 images in each mini-batch. During training, we apply data augmentation by horizontally flipping the images. Note that, during evaluation on a particular dataset, we do not apply any additional CNN training or fine-tuning and dimension reduction.
\subsection{Results and Evaluation}
\label{ssec:res_eval}
Now we evaluate our proposed FR method, called \textit{DeepVisage}, on the most commonly used and challenging facial image datasets based on their specified protocols.

\paragraph{\textit{Labeled Faces in the Wild (LFW) \cite{lfw_huang2014}:}}
\label{par:lfw_eval}
LFW is one of the most popular and challenging databases for evaluating unconstrained FR methods. It consists of 13,233 images of 5,759 identities. It has different evaluation protocols. We follow the \textit{unrestricted-labeled-outside-data} protocol based on the recent trend \cite{learned2015labeled}. The FR task requires verifying 6000 image pairs in 10 folds and report the accuracy. These pairs are equally divided into genuine and impostor pairs and comprises  7.7K images of 4,281 identities.

Table \ref{tab:lfw_comparison} provides the results of our method along with the other state-of-the-art methods. We observe that, our method achieves significant accuracy (99.62\%) and among the top performers, despite the fact that: (a) we use single CNN, whereas Baidu \cite{baiduliu2015} used 10 CNNs to obtain 99.77\% and (b) we train CNN with comparatively much less amount of data and identities, whereas FaceNet \cite{schroff2015facenet} used 200M images of 8M identities to obtain  99.63\%.
\begin{table}[h]
\footnotesize
\centering
\caption{\footnotesize{Comparison of the state-of-the-art methods evaluated on the LFW benchmark \cite{lfw_huang2014}.}}
\label{tab:lfw_comparison}
\begin{tabular}{|c|c|c|c|}
\hline
\textbf{FR method} & \textbf{\begin{tabular}[c]{@{}c@{}}\# of\\ CNNs\end{tabular}} & \textbf{\begin{tabular}[c]{@{}c@{}}Dataset\\ Info\end{tabular}} & \textbf{\begin{tabular}[c]{@{}c@{}}Acc\\ \%\end{tabular}} \\ \hline
\textit{\textbf{DeepVisage (proposed)}} & 1 & 4.48M, 62K & 99.62 \\ \hline
Baidu \cite{baiduliu2015}        & 10 & 1.2M, 1.8K & 99.77 \\ \hline
Baidu \cite{baiduliu2015}        & 1 & 1.2M, 1.8K & 99.13 \\ \hline
FaceNet \cite{schroff2015facenet}        & 1 & 200M, 8M & 99.63 \\ \hline
Sparse ConvNet \cite{sparsifyingsun2015} & 25  & 0.29M, 12K & 99.55 \\ \hline
DeepID3 \cite{sun2015deepid3}            & 25  & 0.29M, 12K & 99.53 \\ \hline
Megvii \cite{zhou2015naive}              & 4 & 5M, 0.2M & 99.50 \\ \hline
LF-CNNs \cite{aifrwen2016latent}     & 25 & 0.7M, 17.2K & 99.50 \\ \hline
DeepID2+ \cite{deepid2psun2015}            & 25  & 0.29M, 12K & 99.47 \\ \hline
Center Loss \cite{centerlosswen2016}     & 1 & 0.7M, 17.2K & 99.28 \\ \hline
MM-DFR \cite{ding2015robust}             & 8 & 0.49M, 10.57K & 99.02 \\ \hline
VGG Face \cite{parkhi2015deep}           & 1 & 2.6M, 2.6K & 98.95 \\ \hline
MFM-CNN \cite{wu2015lightened}         & 1 & 5.1M, 79K & 98.80 \\ \hline
VIPLFaceNet	\cite{viplfacenetliu2016}   & 1 & 0.49M, 10.57K & 98.60 \\ \hline
Webscale \cite{webscaletaigman2015}     & 4 & 4.5M, 55K  & 98.37 \\ \hline
AAL \cite{aalye2016face}				   & 1 & 0.49M, 10.57K & 98.30 \\ \hline
FSS \cite{wang2016face}		   & 9 & 0.49M, 10.57K & 98.20 \\ \hline
Face-Aug-Pose-Syn \cite{masi16dowe} 			& 1 & 2.4M, 10.57K & 98.06 \\ \hline
CASIA-Webface \cite{yi2014learning}      & 1 & 0.49M, 10.57K & 97.73 \\ \hline
Unconstrained FV \cite{chen2016unconstrained} & 1 & 0.49M, 10.5K & 97.45 \\ \hline
Deepface \cite{taigman2014deepface}      & 3 & 4.4M, 4K & 97.35 \\ \hline
\end{tabular}
\end{table}

The results in the Table \ref{tab:lfw_comparison} indicates saturation, because all of the methods achieve close to or more than human performance (97.53\%). Besides, it is argued that matching only 6K pairs is insufficient to justify a method w.r.t. the real world FR scenario \cite{blufrliao2014}. We address these issues by two ways: (a) employ more challenging evaluation metrics and (b) evaluate with the other challenging datasets. 
To this aim, first we follow the BLUFR LFW protocol \cite{blufrliao2014} and measure the true accept rate (TAR) at a low false accept rate (FAR). BLUFR \cite{blufrliao2014} protocol exploits all images of the LFW dataset and evaluates methods based on 10 trials experiments. Each trial computes 47M pair-matching scores (157K positives, 46.9M negatives), which is significantly higher than the 6K scores used in the standard protocol. Within this protocol, we compute the verification rate (VR) at FAR=0.1\% and compare with the methods which reported results\footnote{We do not include results from Baidu \cite{baiduliu2015} (VR@FAR: 99.11\% for single CNN and 99.41\% for 10-CNNs ensembles). The reason is that, we are not sure if they compute results based on the BLUFR protocol \cite{blufrliao2014} or based on the 6K pairs. Note that, we obtain 99.7\% on VR@FAR=0.1\% using the 6K pair-matching scores of the standard protocol.} in this protocol. We observe that: $
\textit{\textbf{DeepVisage \;(proposed)}} \; (98.65) >
Center Loss\footnote{Results computed from the features publicly provided by the authors.}  \; \cite{centerlosswen2016}  \; (92.97\%) >
FSS \; \cite{wang2016face}  \; (89.8\%) >
CASIA \; \cite{yi2014learning}  \; (80.26\%)
$
, i.e., our method obtains the best results published so far. Therefore, this result together with the Table \ref{tab:lfw_comparison} confirm the remarkable performance of \textit{DeepVisage} on the LFW database.
Next, we justify our method by evaluating it on the challenging IJB-A \cite{ijbaKlare2015} dataset. 
\paragraph{\textit{IARPA Janus Benchmark A (IJB-A) \cite{ijbaKlare2015}:}}
\label{par:lfw_eval}
The recently proposed IJB-A database aims at raising the difficulty of FR by incorporating more variations in pose, illumination, expression, resolution and occlusion. It consists of 5,712 images and 2,085 videos of 500 identities. The FR task compares two templates. A template is a set of images and video-frames. The evaluation protocol requires computing the true accept rate (TAR) at a fixed false accept rate (FAR) with various values, e.g., 0.01 and 0.001.

Table \ref{tab:ijb_comparison} presents our results along with the other state-of-the-art methods. We separate the results (with a horizontal line) to distinguish two categories: (1) methods only using a pre-trained CNN; our method belongs to this category and (2) methods use additional learning, such as CNN fine-tuning and metric learning.
From the comparison among the $1^{st}$ category of methods, we observe that, our method provides the best result for FAR at 0.001\% and competitive (second best) at 0.01\%.
By comparing it to the $2^{nd}$ category we observe that, it is also very competitive and provide better results than numerous methods from this category. Besides, similar to \cite{allinoneranjan2016, Sankaranarayanan2016a}, it is possible to exploit our CNN features and further improve the final results with external learning, such as TA \cite{Crosswhite2016}, NAN \cite{nanyang2016} and TPE \cite{Sankaranarayanan2016a}.
\begin{table}[t]
\footnotesize
\centering
\caption{\footnotesize{Comparison of the state-of-the-art methods evaluated on the IJB-A benchmark \cite{ijbaKlare2015}. `-' indicates the information for the entry is unavailable. Methods which incorporates external training (ExTr) or CNN fine-tuning (FT) with IJB-A training data are separated with a horizontal line. VGG-Face result was provided by \cite{Sankaranarayanan2016}. T@F denotes the \textit{True Accept Rate at a	fixed False Accept Rate (TAR@FAR)}.}}
\label{tab:ijb_comparison}
\begin{tabular}{|c|c|c|c|c|}
\hline
\textbf{FR method} & \textbf{\begin{tabular}[c]{@{}c@{}}ExTr\end{tabular}} & \textbf{\begin{tabular}[c]{@{}c@{}}FT\end{tabular}} & \textbf{\begin{tabular}[c]{@{}c@{}}T@F\\0.01\end{tabular}} & \textbf{\begin{tabular}[c]{@{}c@{}}T@F\\0.001\end{tabular}} \\ \hline
\textit{\textbf{DeepVisage (proposed)}}  & N & N & 0.887 & 0.824\\ \hline
VGG Face \cite{parkhi2015deep}           & N & N & 0.805 & 0.604\\ \hline
Face-Aug-Pose-Syn \cite{masi16dowe} 			& N & N & 0.886 & 0.725\\ \hline
Deep Multipose \cite{abdalmageed2016face} & N & N  & 0.787 & - \\ \hline
Pose aware FR \cite{masi2016pose} & N & N  & 0.826 & 0.652 \\ \hline
TPE \cite{Sankaranarayanan2016a} & N & N  & 0.871 & 0.766\\ \hline
All-In-One \cite{allinoneranjan2016} & N & N  & 0.893 & 0.787 \\ \hline \hline \hline
All-In-One \cite{allinoneranjan2016} + TPE & Y & N  & 0.922 & 0.823 \\ \hline
Sparse ConvNet \cite{sparsifyingsun2015} & Y & N  & 0.726 & 0.460\\ \hline
FSS \cite{wang2016face}		   & N & Y  & 0.729 & 0.510\\ \hline
TPE \cite{Sankaranarayanan2016a} & Y & N  & 0.900 & 0.813\\ \hline
Unconstrained FV \cite{chen2016unconstrained} & Y & Y  & 0.838 & -\\ \hline
TSE \cite{Sankaranarayanan2016} & Y & Y  & 0.790 & 0.590\\ \hline
NAN \cite{nanyang2016} & Y & N  & 0.941 & 0.881 \\ \hline
TA \cite{Crosswhite2016} & Y & N  & 0.939 & 0.836 \\ \hline
End-To-End \cite{eteChen2015} & N & Y  & 0.787 & - \\ \hline
\end{tabular}
\end{table}
%
%
%
%
%
\paragraph{\textit{YouTube Faces \cite{ytfwolf2011} (YTF):}}
\label{par:lfw_eval}
The YTF dataset is a widely used FR dataset of unconstrained videos. It consists of 3,425 videos of 1,595 identities. YTF evaluation requires matching 5000 video pairs in 10 folds and report average accuracy. Each fold consists of 500 video pairs and ensures subject-mutually exclusive property.
We follow the \textit{restricted} protocol of YTF, i.e., access to only the similarity information. We report our result
in Table \ref{tab:ytf_comparison}, along with the state-of-the-art methods. Results show that our method provides the best accuracy (96.24\%).

Table \ref{tab:ytf_comparison} also provides the results (separated with a horizontal line) from \textit{unrestricted} protocol, i.e., access to similarity and identity information of the test data. 
We observe that our method is very competitive to the best accuracy, although it follows the \textit{restricted} protocol. The VGG Face \cite{parkhi2015deep} provides results with both protocols and shows that accuracy increases significantly (from \textit{restricted}-91.6\% to \textit{unrestricted}-97.3\%) when they learn their CNN feature embedding using the YTF training data. Based on this observation, we can predict that our result (96.24\%) can be further enhanced by training or fine tuning with the YTF data.
\begin{table}[h]
\footnotesize
\centering
\caption{\footnotesize{Comparison of the state-of-the-art methods evaluated on the Youtube Face \cite{ytfwolf2011}. \textit{\textbf{Ad.Tr.}} denotes additional training is used.}}
\label{tab:ytf_comparison}
\begin{tabular}{|c|c|c|}
\hline
\textbf{FR method} & \textbf{\begin{tabular}[c]{@{}c@{}}Ad.Tr.\end{tabular}} & \textbf{\begin{tabular}[c]{@{}c@{}}Accuracy (\%)\end{tabular}} \\ \hline
\textit{\textbf{DeepVisage (proposed)}}  & N & 96.24\\ \hline
VGG Face \cite{parkhi2015deep}           & N & 91.60\\ \hline
Sparse ConvNet \cite{sparsifyingsun2015} & N & 93.50\\ \hline
FaceNet \cite{schroff2015facenet}        & N & 95.18 \\ \hline
DeepID2+ \cite{deepid2psun2015}          & N & 93.20 \\ \hline
Center Loss \cite{centerlosswen2016}     & N & 94.90 \\ \hline
MFM-CNN \cite{wu2015lightened}         & N & 93.40 \\ \hline
CASIA-Webface \cite{yi2014learning}      & Y & 92.24 \\ \hline
Deepface \cite{taigman2014deepface}      & Y & 91.40 \\ \hline
\hline \hline
VGG Face \cite{parkhi2015deep}           & Y & 97.30\\ \hline
NAN \cite{nanyang2016} 				    & Y  & 95.72 \\ \hline
\end{tabular}
\end{table}
\paragraph{\textit{Cross-Age Celebrity Dataset (CACD) \cite{chen2015facecacd}:}}
\label{par:lfw_eval}
CACD is a recently released dataset, which aims to ensure large variations of the ages in the wild. It consists of 163,446 images of 2000 identities with the age range from 16 to 62. CACD evaluation requires verifying 4000 image pairs in ten folds and report average accuracy. Table \ref{tab:cacd_comparison} reports the results of \textit{DeepVisage} along with the state-of-the-art methods. It shows that our method provides the best accuracy. Moreover, it is better than LF-CNN \cite{aifrwen2016latent}, which is a recent method specialized on age invariant face recognition.

\begin{table}[h]
\footnotesize
\centering
\caption{\footnotesize{Comparison of the state-of-the-art methods evaluated on the CACD \cite{chen2015facecacd} dataset. VGG \cite{parkhi2015deep} result is obtained from \cite{wu2015lightened}.}}
\label{tab:cacd_comparison}
\begin{tabular}{|c|c|c|}
\hline
\textbf{FR method} & \textbf{\begin{tabular}[c]{@{}c@{}}Accuracy (\%)\end{tabular}} \\ \hline
\textit{\textbf{DeepVisage (proposed)}}  & 99.13\\ \hline
LF-CNNs \cite{aifrwen2016latent}         & 98.50 \\ \hline
MFM-CNN \cite{wu2015lightened}           & 97.95 \\ \hline
VGG Face \cite{parkhi2015deep}           & 96.00\\ \hline
CARC \cite{chen2015facecacd}             & 87.60 \\ \hline \hline
Human, Avg.             				   & 85.70 \\ \hline
Human, Voting \cite{chen2015facecacd}    & 94.20 \\ \hline
\end{tabular}
\end{table}
The evaluations of \textit{DeepVisage} (proposed method) across different challenging datasets prove that it not only achieves significant performance but also generalizes 
very well. It overcomes several of the difficulties which make unconstrained FR a challenging task.
\subsection{Analysis and Discussion}
\label{ssec:disscussion}
We perform further analysis to highlight the influences of several aspects, such as: (a) training datasets; (b) CNN models and depth; (c) normalization and (d) activation functions. Therefore, we modify and train our CNN model and observe the accuracy and TAR@FAR=0.01 on LFW. Table \ref{tab:analysis_discussion_db} presents the results.

First, we study the influence of training the proposed CNN with different datasets. It helps us to understand the capacity of the CNN to learn facial representation and identify the requirements to achieve better performance. 
The top part of Table \ref{tab:analysis_discussion_db} presents the analysis w.r.t. different datasets, from which we observe that: (a) CNN performance increased by training with larger number of images as well as identities, 
the best results are obtained with the largest dataset, i.e., MSCeleb \cite{mscelebguo16}; 
(b) synthesized images help to enhance performance, we see this from the pose augmented CASIA \cite{yi2014learning, masi16dowe} dataset; (c) a dataset with more variations per identity helps even with a relatively lower number of images and identities, we see this by comparing the CASIA \cite{yi2014learning} and UMD \cite{umdfacesbansal2016} datasets; and (d) large number of images with smaller number of identities may not help, we see this from the VGG Face \cite{parkhi2015deep} dataset. Besides, we analyze the dataset uniformity or balance issue, i.e.,  number of images-per-identity, see bottom part of of Table \ref{tab:analysis_discussion_db}. We use the MSCeleb \cite{mscelebguo16} dataset for this experiment. We see that, while maintaining certain balance is necessary, it is equality important to train CNN with a larger dataset. We obtain the best performance by keeping only the identities with 30 images or more.

\begin{table}[t]
\footnotesize
\centering
\caption{\footnotesize{Analysis of the influences from training databases, size and number of classes. T@F denotes the \textit{True Accept Rate at a fixed False Accept Rate (TAR@FAR)}.}}
\label{tab:analysis_discussion_db}
\begin{tabular}{|c|c|c|c|c|}
\hline
\textbf{Aspect} & \textbf{Add. info} & \textbf{\begin{tabular}[c]{@{}c@{}}Acc\\\%\end{tabular}} & \textbf{\begin{tabular}[c]{@{}c@{}}T@F\\0.01\end{tabular}}\\ \hline \hline
\hline
\textit{\textbf{DB}} & \textbf{\textit{Size, Class}} &  &\\ \hline
CASIA \cite{yi2014learning} & 0.43M, 10.6K & 99.00 & 0.988 \\ \hline
Pose-CASIA \cite{masi16dowe} & 1.26M, 10.6K & 99.15 & 0.992 \\ \hline
UMDFaces \cite{umdfacesbansal2016} & 0.34M, 8.5K & 99.15 & 0.992 \\ \hline
VGG Face \cite{parkhi2015deep} & 1.6M, 2.6K & 98.40 & 0.975 \\ \hline
MSCeleb \cite{mscelebguo16} & 4.2M, 62.5K & 99.62 & 0.997 \\ \hline
\hline
\hline
\textit{\textbf{Min samp/id}} & \textbf{\textit{Size, Class}} &  &\\ \hline
10 & 4.48M, 62.7K & 99.56 & 0.996 \\ \hline
30 & 4.47M, 62.5K & 99.62 & 0.997 \\ \hline
50 & 3.91M, 47.3K & 99.60 & 0.997 \\ \hline
70 & 3.11M, 33K & 99.55 & 0.996 \\ \hline
100 & 1.5M, 12.7K & 99.23 & 0.991 \\ \hline
\end{tabular}
\end{table}

\begin{figure}[t]
\centering
\includegraphics[scale=0.4]{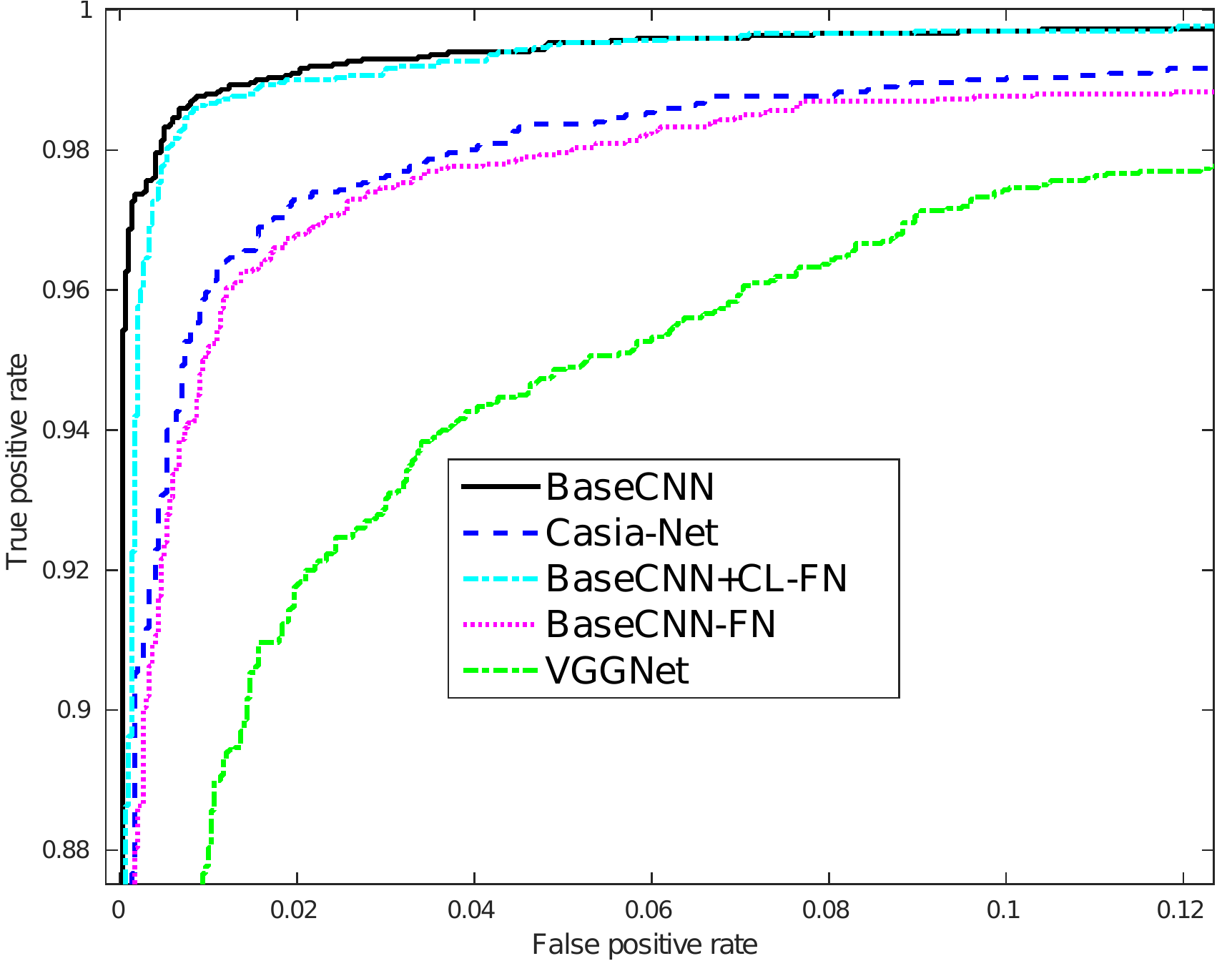}
\caption{\footnotesize{Illustration of the ROC plots for different CNN models evaluates on the LFW \cite{lfw_huang2014} dataset.}}
\label{fig:roc_det_plot}
\end{figure}

Next, we analyze the results based on different CNN components and models. Table \ref{tab:analysis_discussion_cnn} and Fig. \ref{fig:roc_det_plot} present the results with different forms, where we train all settings with the CASIA \cite{yi2014learning} dataset. Our observations are: (a) the proposed CNN model obtains better performance by including feature normalization (FN) before loss computation, we see this by comparing with the center loss \cite{centerlosswen2016} and without FN based results and (b) it obtains better accuracy than the other commonly used CNNs (for FR), such as the VGG-Net \cite{parkhi2015deep} and CASIA-Net \cite{yi2014learning}. Note that, we do not directly compare with other loss functions (within our CNN model) as the center loss \cite{centerlosswen2016} has been shown to be more efficient than those.  Additionally, we trained our CNN with ReLU instead of PReLU and observe that it decreases accuracy by approximately 0.5\%. 
In terms of complexity (measured with the number of parameters in Table \ref{tab:analysis_discussion_cnn}), our model is more complex than the simpler models (Cas-Net and CN-mod). However, it is much simpler than the VGG-Net \cite{parkhi2015deep}. Results indicate that, while a simpler model may limit\footnote{We train the CN-mod (see Table \ref{tab:analysis_discussion_cnn}) with the MSCeleb dataset and observed that, compared to our proposed CNN model CN-mod provides lower results and generalizes poorly.} the FR performance, a complex model is prone to overfitting. Perhaps this is the reason why the VGG-Net \cite{parkhi2015deep} requires additional fine-tuning on the target datasets.
The above analyses justify the efficiency of our proposed CNN model.
\begin{table}[h]
\footnotesize
\centering
\caption{\footnotesize{Study the influences from CNN related issues. All CNN models are trained with the CASIA \cite{yi2014learning} dataset. CL- center loss \cite{centerlosswen2016}, FN- feature normalization. \textit{\textbf{CN-mod}} modifies the \textit{\textbf{Cas-Net}} \cite{yi2014learning} by replacing \textit{Pool} layer with a \textit{FC} layer of 512 neurons.}}
\label{tab:analysis_discussion_cnn}
\begin{tabular}{|c|c|c|c|}
\hline
\textbf{Settings} & \textbf{\textit{\# params}} & \textbf{\begin{tabular}[c]{@{}c@{}}Acc\\ \%\end{tabular}} & \textbf{\begin{tabular}[c]{@{}c@{}}T@F\\0.01\end{tabular}}\\
\hline
Base-CNN (\textit{\textbf{proposed}}) & 40.5M & 99.00 & 0.988 \\ \hline
Base-CNN - FN & 40.5M & 97.40 & 0.954 \\ \hline
Base-CNN + CL - FN & 44.8M & 98.85 & 0.986 \\ \hline
VGG-Net \cite{parkhi2015deep} & 182M & 95.15 & 0.883 \\ \hline
Cas-Net \cite{yi2014learning} & 6M & 97.10 & 0.938 \\ \hline
CN-mod & 8M & 97.50 & 0.956 \\ \hline
\end{tabular}
\end{table}

\begin{figure}[h]
\centering
\includegraphics[scale=0.23]{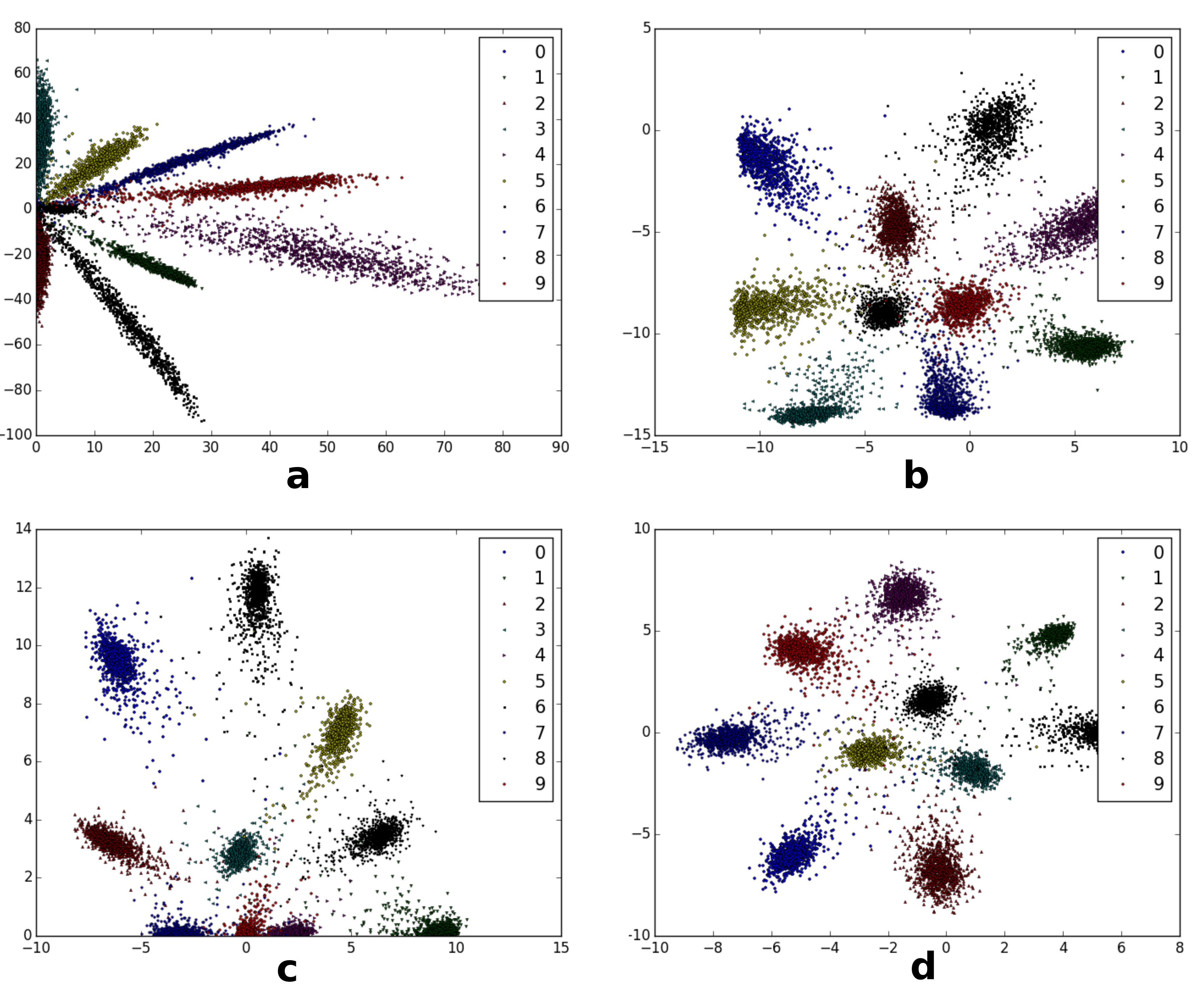}
\caption{\footnotesize{2D visualization of the MNIST \cite{lecun1998gradient} digits features, which are obtained by using same baseline CNN model and training settings. CL \cite{centerlosswen2016} parameters are set to $\lambda=0.003$ and $\alpha=0.5$. \textbf{a.} CNN without FN and CL; \textbf{b.} CNN with FN; \textbf{c.} CNN with CL; and \textbf{d.} CNN with FN and CL.}}
\label{fig:comp_bn_cl}
\end{figure}

We observe that, feature normalization (FN) before the loss computation plays a significant role in the performance. In order to gain further insights, we conduct experiments and visualize the features of the MNIST digits in the 2D space. This is similar to the visualization recently shown in \cite{centerlosswen2016} and hence we also provide a comparison with the center loss (CL). The CNN is composed of 6 convolution, 2 pool and 1 FC (with 2 neurons for 2D visualization) layers. We optimize it using the softmax loss. Fig. \ref{fig:comp_bn_cl} provides the illustration, from which we observe that: (a) FN provides a better feature discrimination in the normalized 2D space, see Fig. \ref{fig:comp_bn_cl}-b; (b) CL enforces the features towards its representative center and hence shows discrimination, see Fig. \ref{fig:comp_bn_cl}-c and (c) CL+FN does not provide much additional discrimination, see Fig. \ref{fig:comp_bn_cl}-b and Fig. \ref{fig:comp_bn_cl}-d. 
These observations reveal that, by exploiting the FN appropriately we can ensure feature discrimination and hence no additional loss function, e.g., CL, is necessary.

Finally, we investigate the incorrect results by observing the face image pairs in which \textit{DeepVisage} failed. \textit{Appendix} \ref{sec:error_analysis} provides the illustrations of the false accept/reject cases from the different datasets. We observe that, on LFW it failed (11/20 error cases) when the eyes are occluded by glasses or a cap. Incorrect CACD results and higher false rejection rate indicate that our method (although provides best accuracy) encounters difficulties to recognize the same person from the images of different ages. Incorrect results from YTF often suffers from high pose and perhaps low image resolution. IJB-A results reveal that our method needs to take care of the face images with extreme pose variations. Indeed, during the IJB-A experiments, we are forced to keep a large number of images as un-normalized due to the failure of landmarks detection for them. Based on empirical evidences, we believe that these un-normalized faces cause the degradation of our performance.
Besides, the results from YTF and IJB-A indicate that we may need to use a better distance computation strategy.

\section{Conclusion}
\label{sec:conclusion}
In this paper we present a single-CNN based FR method which achieves state-of-the-art performance and exhibits excellent ability of generalize across different FR datasets. Our method, called \textit{DeepVisage}, performs face verification based on a given pair of single images, templates and videos. It consists in a deep CNN model which is simple and straightforward to train. Overall, \textit{DeepVisage} is very easy to implement, thanks to the residual learning framework, feature normalization, softmax loss and the simplest distance. It successfully demonstrates that, in order to achieve state-of-the-art results it is not necessary to develop a complicated FR method by using complex training data preparation and CNN learning procedure. We foresee several future perspectives of this work, such as: (a) train CNN with a larger and more balanced dataset, which can be constructed by combining multiple publicly available datasets or by adopting the face synthesizing strategy \cite{masi16dowe} with the existing one; (b) enhance FR performance by incorporating failure detection based technique \cite{steger2016failure}, particularly for face and landmarks detection and (c) incorporate better distance computation method for the template and video comparison, e.g., use softmax based distance \cite{masi16dowe}.

%
%
\appendix
\section{Analysis of the incorrect results}
\label{sec:error_analysis}
In this section, we provide examples of the incorrect results observed from the face verification experiments on different datasets.

\begin{figure}[h]
\centering
\subfloat[]{\includegraphics[scale=0.47]{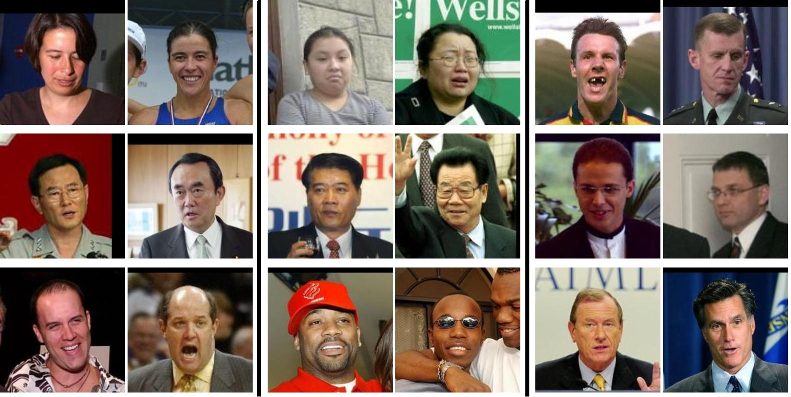}}
\hfil
\subfloat[]{\includegraphics[scale=0.47]{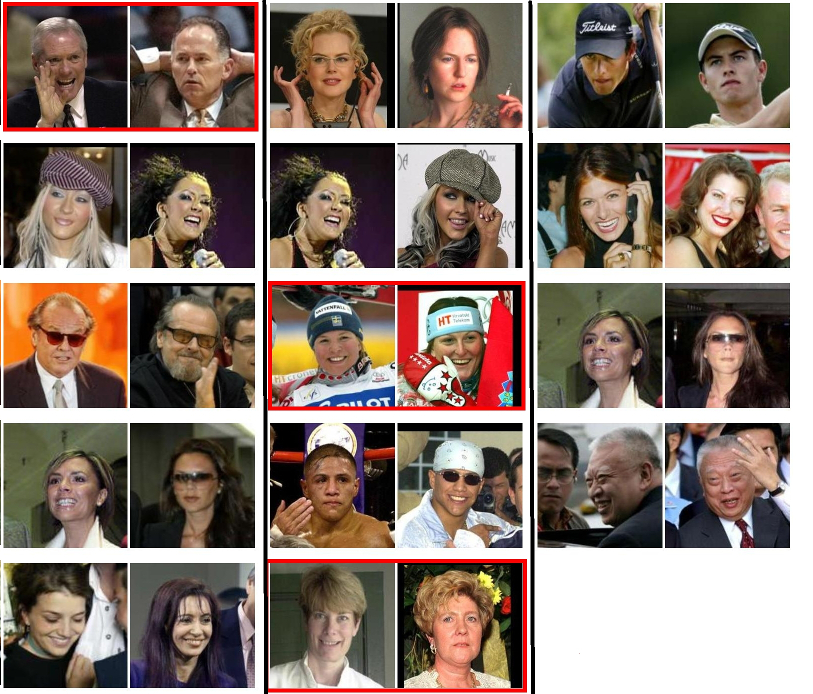}}
\caption{Illustration of the false accepted/rejected image pairs from the LFW \cite{lfw_huang2014} benchmark. (a) false accepted pairs and (b) false rejected pairs. The red colored rectangles indicate the examples which were erroneously labeled in the dataset.}
\label{fig:lfw_error}
\end{figure}

\paragraph{LFW \cite{lfw_huang2014}:}
Figure \ref{fig:lfw_error} provides the examples of the failure cases on the LFW \cite{lfw_huang2014} benchmark. The ratio of false accept vs reject is \textbf{\textit{1:1.56}}. Note that our method achieves 99.62\% accuracy on LFW. In Figure \ref{fig:lfw_error}(b) three pairs are marked with red colored rectangles. These pairs are erroneously labeled in the dataset, which means our method makes correct judgment on them and hence the accuracy further increases to 99.67\% by considering them as correct match.

\begin{figure}[h]
\centering
\subfloat[]{\includegraphics[scale=0.47]{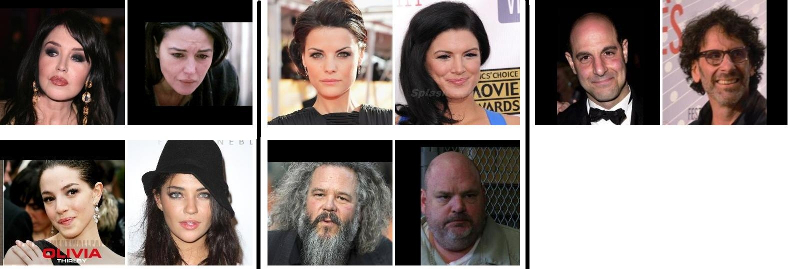}}
\hfil
\subfloat[]{\includegraphics[scale=0.47]{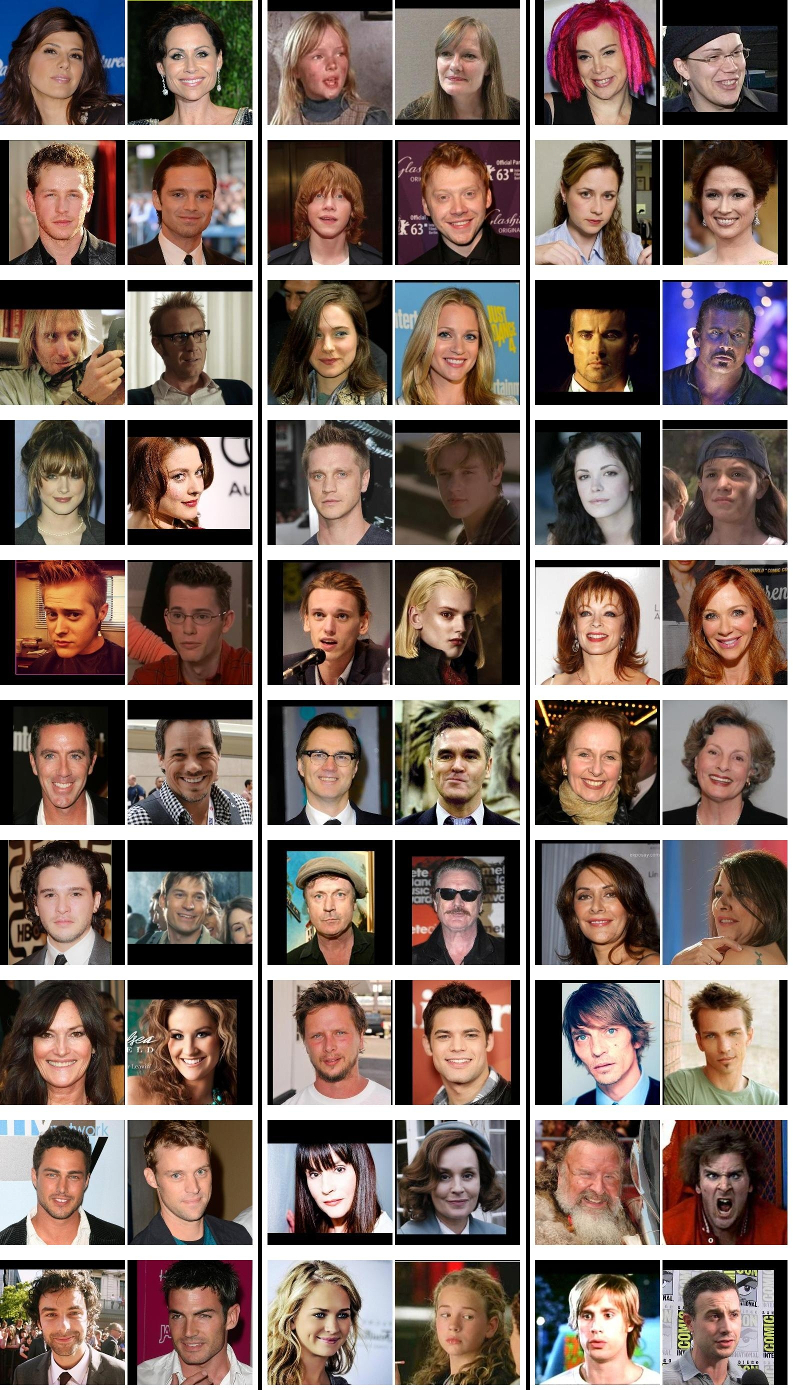}}
\caption{Illustration of the false accepted/rejected image pairs from the CACD-VS \cite{chen2015facecacd} dataset. (a) false accepted pairs and (b) false rejected pairs.}
\label{fig:cacd_error}
\end{figure}

\begin{figure*}[h]
\centering
\subfloat[]{\includegraphics[scale=0.4]{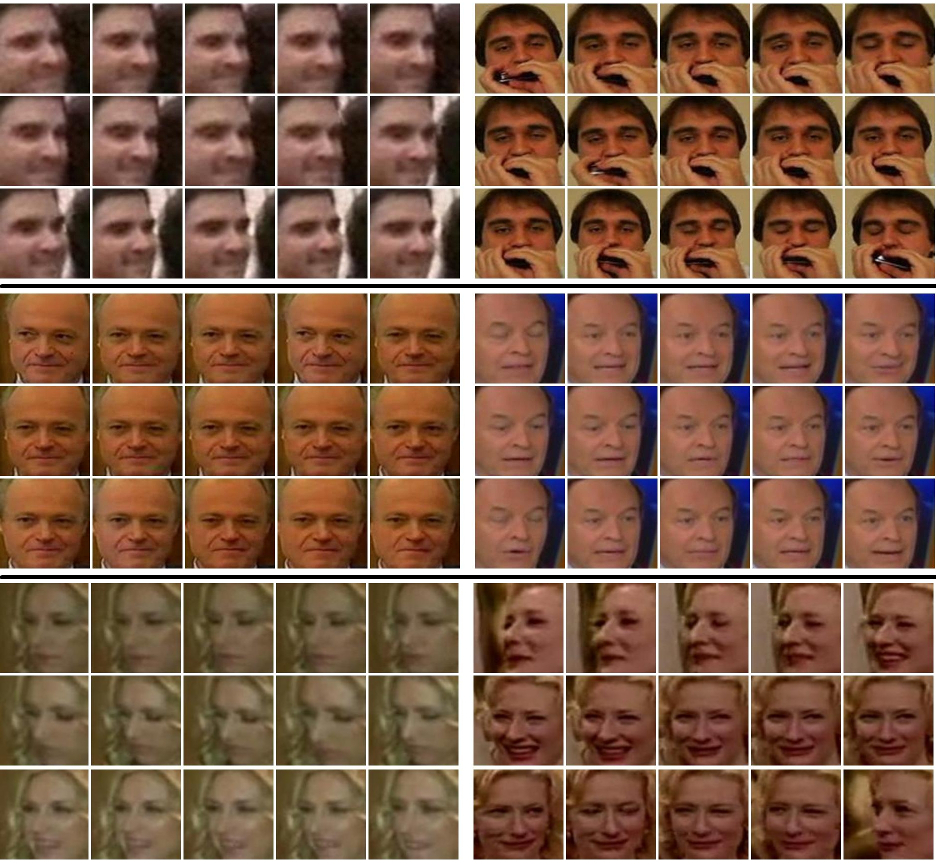}}
\hfil
\subfloat[]{\includegraphics[scale=0.4]{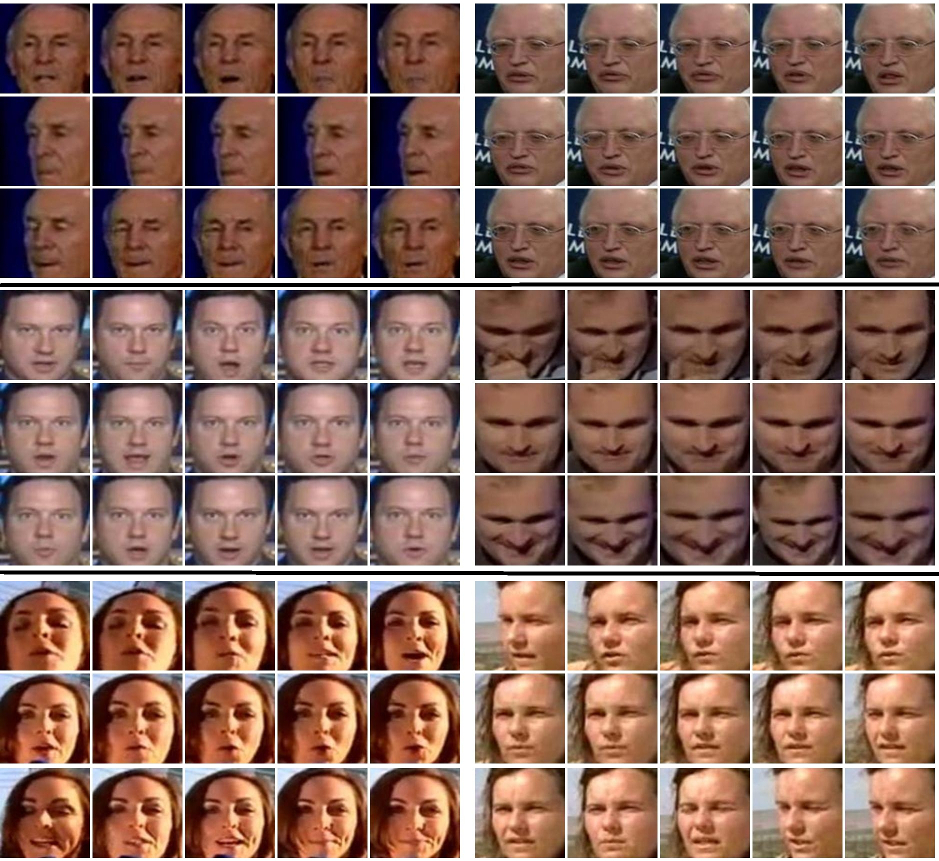}}
\caption{Illustration of the false accepted/rejected video pairs from the YTF \cite{ytfwolf2011} dataset. (a) false accepted pairs and (b) false rejected pairs.}
\label{fig:ytf_error}
\end{figure*}

\begin{figure*}[h]
\centering
\subfloat[]{\includegraphics[scale=0.8]{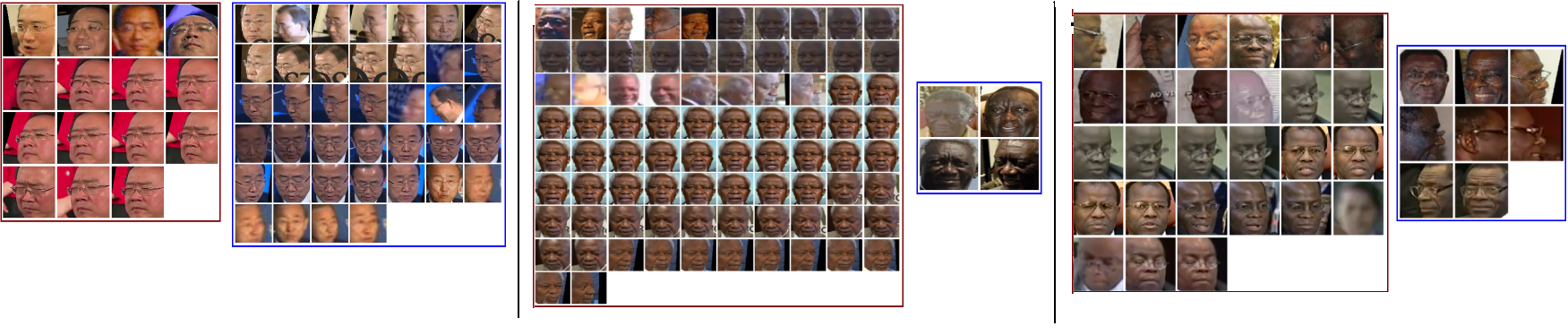}}
\hfil
\subfloat[]{\includegraphics[scale=1]{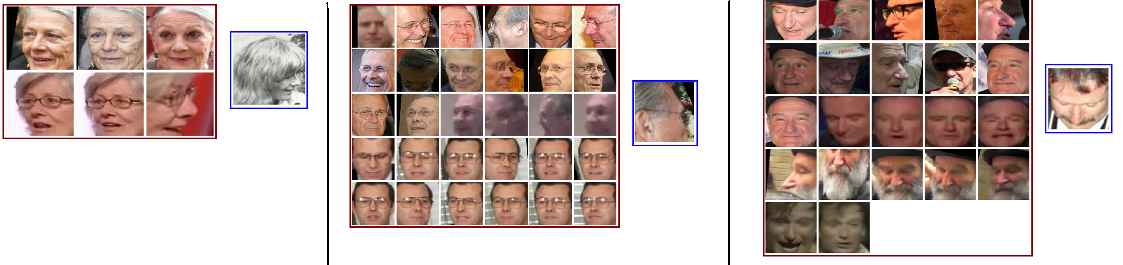}}
\caption{Illustration of the false accepted/rejected template pairs from the IJB-A \cite{ijbaKlare2015} dataset. (a) false accepted pairs and (b) false rejected pairs.}
\label{fig:ijba_error}
\end{figure*}

\paragraph{CACD-VS \cite{chen2015facecacd}:}
Figure \ref{fig:cacd_error} provides the examples of the failure cases on the CACD \cite{lfw_huang2014} dataset. The ratio of false accept vs reject is \textbf{\textit{1:6}}.
\paragraph{YTF \cite{ytfwolf2011}:}
Figure \ref{fig:ytf_error} provides few examples of the failure cases on the YTF \cite{ytfwolf2011} dataset. The ratio of false accept vs reject is \textbf{\textit{1:2.2}}. In Figure \ref{fig:ytf_error}, we only show top three mistakes (sorted based on their similarity score) in terms of false accept and reject.
\paragraph{IJB-A \cite{ijbaKlare2015}:}
Figure \ref{fig:ijba_error} provides few examples of the IJB-A  \cite{ijbaKlare2015} failure cases. The ratio of false accept vs reject is \textbf{\textit{1:5.15}}. In Figure \ref{fig:ijba_error}, we only show top three mistakes (sorted based on their similarity score). From the falsely rejected template pairs, we observe that: (a) one pair has only one image in the template; (b) the pre-processor fails to detect face as well as landmarks and (c) the images in the template have very high pose and large occlusion which causes important face attributes to be absent.
%

{\small
\bibliographystyle{ieee}
\bibliography{dv}
}

\end{document}